# 3D ARCHITECT: AN AUTOMATED APPROACH TO THREE DIMENSIONAL MODELING

**[1]SUNIL TIWARI, [2]PAYAL FOFADIYA, [3]VICKY VISHWAKARMA**

[1,2,3]Information Technology, V.E.S. Institute of Technology, Chembur, Mumbai, India
E-mail: sunil28071987@gmail.com, fofadiyapayal@gmail.com, mailforvicky@gmail.com

**Abstract**- The aim of our paper is to render an object in 3-dimension using a set of its orthographic views. Corner detector (Harris Detector) is applied on the input views to obtain control points. These control points are projected perpendicular to respective views, in order to construct an envelope. A set of points describing the object in 3-dimension, are obtained from the intersection of these mutually perpendicular envelopes. These set of points are used to regenerate the surfaces of the object using computational geometry. At the end, the object in 3-dimension is rendered using OpenGL.

## I. INTRODUCTION

There is growing interest in transforming images to models, i.e., to construct geometric and descriptive models of 3D objects from sensed data for various applications. Accurate models of existing free-form objects are required in many emerging applications such as object animation, visualization in virtual museums and vision augmented environments. The difficult part in generating 3D models is to define a technique to acquire data in order to represent the object in 3 dimensions. In this paper we have described a technique to acquire this data from a set of images of the object. This set of images consists of orthographic views of the object taken from various angles (front view, top view, side view). Finally, the 3D points obtained from our method are combined with the work of Nina Amenta et al. to generate the surfaces of the object. The paper is organized as follows. Section III describes control point detection that is used in subsequent sections. Envelope creation and detection of 3D points is studied in Section IV. Surface reconstruction is established in section V. Finally some experimental results are provided in section VI.

## II. RELATED WORK

The idea of generating the 3 dimensional models is not new. The work of the surveying technique of photogrammetry of Fei Dai et al. takes a completely different approach by deriving metric information about an object through measurements conducted on photographs of the object. The work of Jiang Yu Zheng et al. reconstructs a 3D graphics model of an object with specular surfaces by its rotation. Continuous images are taken to measure highlights on the smooth surfaces and their motion. The work of Chitra Dorai et al. uses laser range scanner to acquire the data. It is indeed a robust method and has no loss of information of the object surfaces. However, this method is quite expensive and involves a lot of precision to be taken into account.

## III. CONTROL POINT DETECTION

2-D engineering drawings are still used widely and play essential role in traditional engineering, because they are the primary format of technical presentation and communication in design and manufacturing. In our work we have used images, Fig.(1), that represents the orthographic views of the object under consideration.

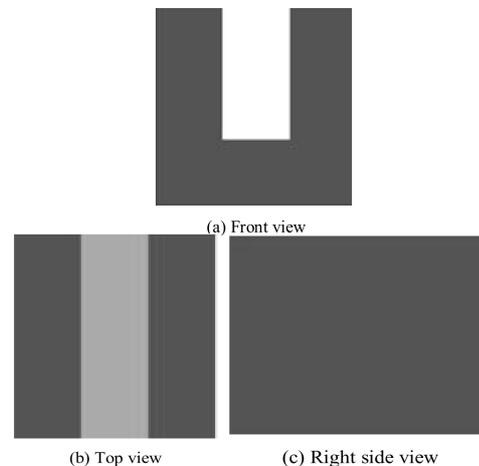

(a) Front view

(b) Top view   (c) Right side view
**Fig.(1) Orthographic view of the object taken from front, top and side**

We have chosen to follow the 1st angle convention, Fig.(2), to represent orthographic views and for further descriptions throughout our paper.

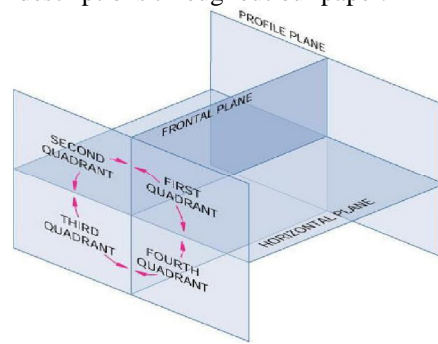

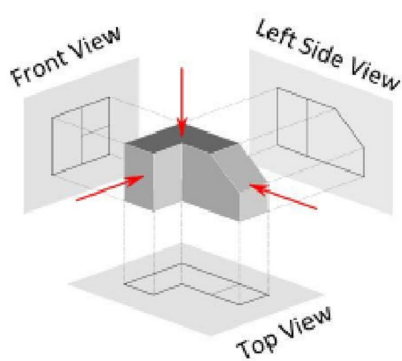

Fig.(2) Convention of 1st angle orthographic projection

Control points are defined as points of strong curvature in an image. For example, one can consider the points making the edge of an object (a photometric corner). There are many established methods to find points from images like Kitchen-Rosen field operator, Moravec operator, Susan detector, Plessey (Harris) detector, etc. We choose Harris (Plessey) corner detector because it is very robust with respect to image noise and gives a good localization of the control points, Fig.(3).

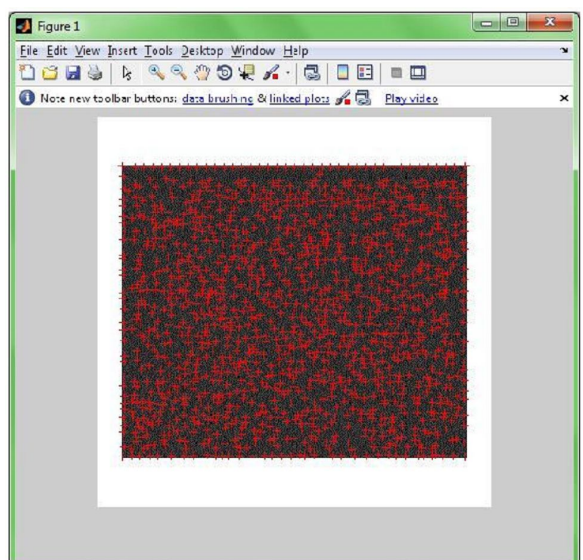

Fig.(3) Control points, in red, are detected using Harris Detector

The formulation of Harris detector involves a cornerness measurement function. Cornerness measurement function, R is defined as follows:

$$R = det(M) - 0.04 \times trace(M)^2$$

where;

$$M = \begin{bmatrix} L_x^2 & L_x L_y \\ L_x L_y & L_y^2 \end{bmatrix}$$

$$L_x = G_{xy} * I_x$$

$$L_y = G_{xy} * I_y$$

where $*$ is convolution operator.

$$G_{xy} = \frac{1}{2\pi\sigma^2} e^{\frac{-(x^2+y^2)}{2\sigma^2}}$$

and $I_x$, $I_y$ are partial derivatives of the image along width x and height y respectively.

## IV.  ENVELOPE CREATION AND DETECTION OF 3D POINTS

The control points obtained in the previous step are used in envelope creation. The process of envelope creation requires an understanding of projections. Projections of a view are lines drawn from control points and perpendicular to the view. These projection lines drawn from each of the control points form a closed envelope describing the view. Fig.(4), shows the closed envelopes for each of the views.

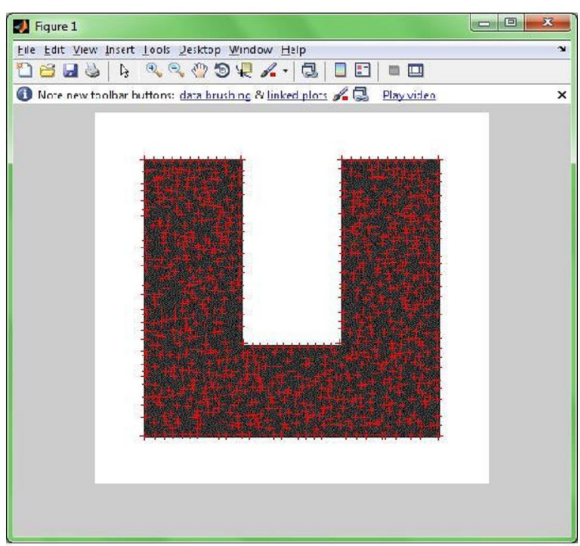

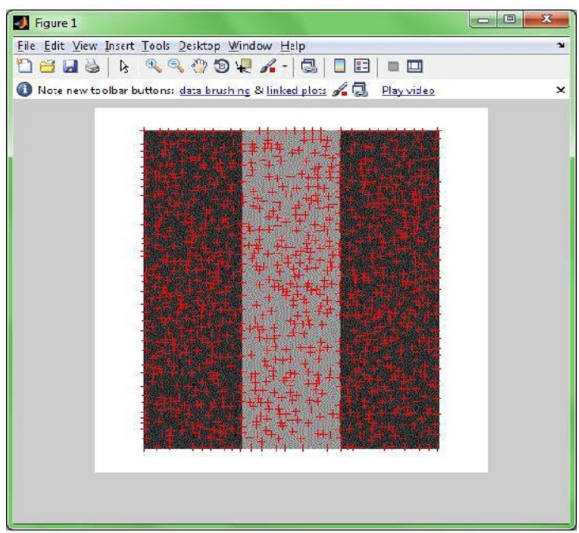

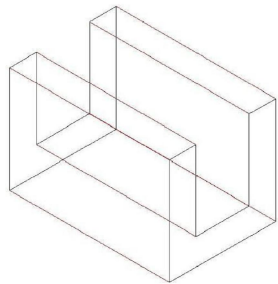

(a) Front view envelope

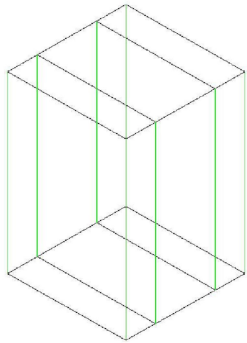

(b) Top view envelope

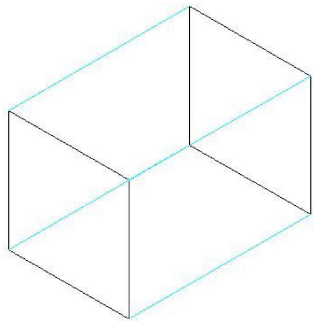

(c) Left side view envelope

**Fig.(4) Schematic representation of envelope creation**

The envelopes created in the previous step are further used for determining the points in 3D. The points of intersection of any two mutually perpendicular envelopes, give a set of points in 3D. This set of points describes the volume of the object, Fig.(5).

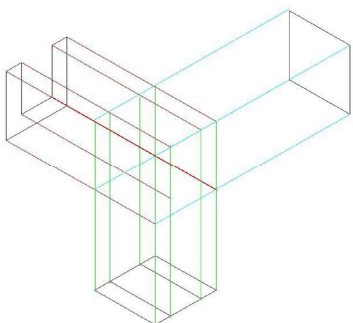

**Fig.(5) Intersection of envelopes**

In order to determine this set of points describing the object in 3D, we require two or more mutually perpendicular envelopes. Two mutually perpendicular envelopes yield sufficient number of points to describe the object in three dimensions. More complex objects require more than two mutually perpendicular envelopes to describe the object sufficiently.

## V. SURFACE CONTRUCTION

The 3D points obtained from the intersection of the envelopes are combined with the work of Nina Amenta et al. to generate the surfaces of the object. Nina Amenta et al. used Delaunay triangulation along with Voronoi diagrams to reconstruct the surfaces from a cloud of three dimensional points. "Crust Theorem" devised in for surface reconstruction is as detailed below.

1. Compute Voronoi diagram of the set of points in „S‟.

2. For each sample point s

2a. Compute vertex $p^+$ of cell Vor(s) farthest from s.

2b. Compute vertex $p^-$ of cell Vor(s) farthest from such that $dot(sp^+, sp^-) < 0$, where dot() gives dot product of two vectors.

3. Let P denote all poles $p^+$ and $p^-$.

4. Compute Delaunay triangulation of union(S, P) where union() gives union of two sets.

5. Extract triangles for which all 3 vertices are in S.

Fig.(6) shows the output of surface reconstruction.

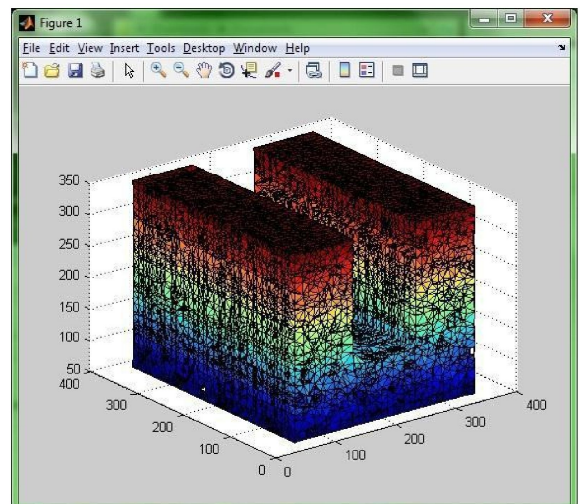

**Fig.(6) Surface construction using Crust theorem**

## VI. IMPLEMENTATION

Time has always been a concern when implementing an algorithm which requires computational geometry evaluations. Our multi-threaded implementation ensures that the processing time of the system is allowable. Hence, the technique gives a perfect trade-off between model accuracy and processing time. We have used the in- build routines of MATLAB for image processing and obtaining Delaunay Triangulation of sample points. Some of the experimental results obtained are shown below in, Table 1. Table 1, indicates the number of points detected in envelope intersection, number of triangles obtained in Delaunay Triangulation and total processing time of the model. We have used OpenGL to render the model for various other manipulations like scaling and rotation.

| Model No. | No. of 3D points | No. of triangles | Time (min.) |
|---|---|---|---|
| Model1 | 42,401 | 2,46,631 | 1.378 |
| Model2 | 15,322 | 94,223 | 0.660 |
| Model3 | 20067 | 1,17,707 | 0.647 |

Table 1

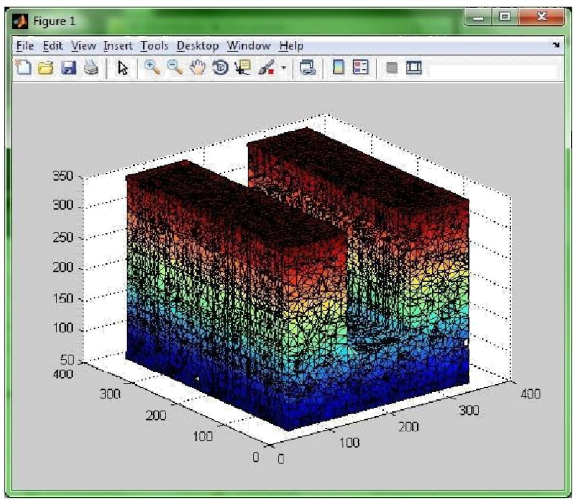

Model1

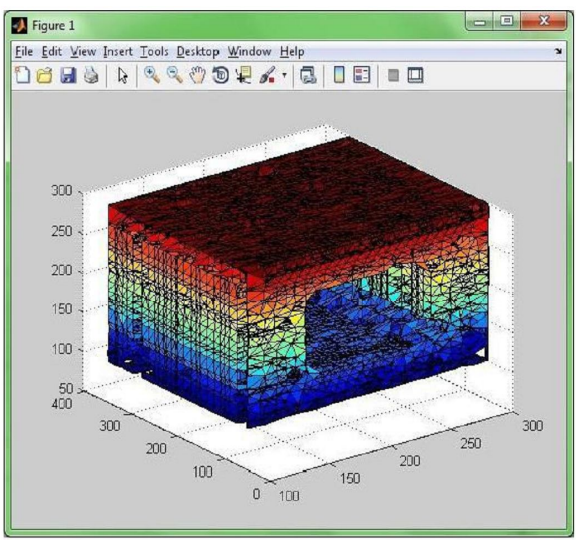

Model2

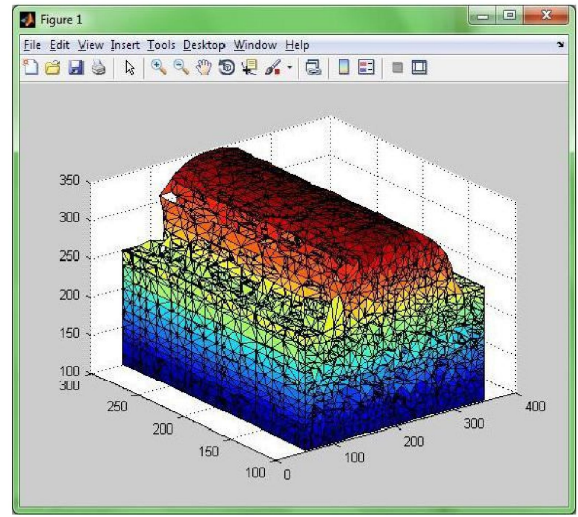

Model3

**CONCLUSION**

The technique presented above provides an advantage of automation. It is a simple, efficient and practically feasible approach to convert 2D images into 3D models. Our approach requires no human intervention for this conversion. Our work has a scope in various fields like architectural design, component inspection, virtual museums display, entertainment industries, etc. Machine parts are more structured in nature; hence our work can automate the process of their component inspection. Dynamic image sequences can be combined with computer animation and our work, to animate the dynamic changes an object undergoes over the time in terms of shape and orientation.

★★★